\documentclass[runningheads]{llncs}
\usepackage{graphicx}
\usepackage{enumerate}
\usepackage{cleveref}
\begin{document}
\title{Dear XAI Community, We Need to Talk! 
\newline
\small{Fundamental Misconceptions in Current XAI Research}
}
%
\titlerunning{Dear XAI Community, We Need to Talk!}
%
\author{Timo Freiesleben \inst{1} \and Gunnar König \inst{2}}

%
\authorrunning{Freiesleben and König}
%

\institute{Cluster of Excellence ``Machine Learning in Science'', Tübingen University, Germany, \email{timo.freiesleben@uni-tuebingen.de} \and Department of Statistics, LMU Munich, Germany, \email{gunnar.koenig@stat.uni-muenchen.de}}

\maketitle              
\begin{abstract}
Despite progress in the field, significant parts of current XAI research are still not on solid conceptual, ethical, or methodological grounds. Unfortunately, these unfounded parts are not on the decline but continue to grow. Many explanation techniques are still proposed without clarifying their purpose. Instead, they are advertised with ever more fancy-looking heatmaps or only seemingly relevant benchmarks. Moreover, explanation techniques are motivated with questionable goals, such as building trust, or rely on strong assumptions about the ’concepts’ that deep learning algorithms learn. In this paper, we highlight and discuss these and other misconceptions in current XAI research. We also suggest steps to make XAI a more substantive area of research.

\end{abstract}
\section{Introduction}
This is an unusual paper from start to end. We don't start the paper with generic examples of great Machine Learning (ML) achievements. The thoughts in this paper are directed at people who are already working on eXplainable Artificial Intelligence (XAI), so we are long past promotional talks. Our goals with this paper are twofold: 1. to highlight misconceptions within parts of the XAI community in past and current research; 2. to provide constructive feedback and steps forward to make XAI a scientific discipline that actually improves ML transparency. 

After wrapping our heads around XAI-related topics for a couple of years, we became increasingly frustrated whenever we attended a workshop or conference on the topic. We do not claim that no progress is being made or that no high-quality research is being conducted. However, we are saddened that many computational, intellectual, and financial resources are being poured into projects that, in our view, do not stand on solid grounds: 

\begin{itemize}
    \item proposals for new interpretation techniques that serve no clear purpose
     \item anecdotal evidence from intuitive-looking heatmaps or "benchmarks" on seemingly relevant criteria are used as a substitute for a clear motivation 
    \item explanations are generated that mislead humans into trusting ML models without the models being trustworthy 
\end{itemize}
Instead of swallowing our frustration, we decided to canalize it into this paper with the hope of helping researchers avoid such projects that might be technically interesting but conceptionally unfounded. We believe that such a debate is especially urgent since funding for XAI research is inexorably high, and the community is ever-growing. Without clear purposes and proper conceptual foundations, the XAI boom could lead to a bubble endangered to implode. We would like to see our field become a pillar of ML transparency rather than the ML trust-washing machine.

The perspective we will take is more of a philosophical bird's eye view of XAI research. It is not our style to expose specific papers by pointing out their flaws. We also feel that this is not necessary because the misconceptions discussed are 'elephants in the room' in our community. Before sharing our thoughts, we would like to point the reader to work that guided our perspective on XAI and that may help to underpin our arguments.

\section{Related Work}
Many papers criticize XAI on various grounds, and we believe many of the criticisms still apply to current XAI. We focus on the critiques that most impacted the community and/or our thoughts.

In his seminal paper, Zachary Lipton argues that XAI lacks a proper problem formulation and that this problem must be tackled to make progress as a field \cite{lipton2018mythos}. Instead of a well-defined goal, XAI offers a potpourri of motivations for explainability, such as increasing trust, fairness, or understanding. Summarized, he argues that: ``When we have solid problem formulations, flaws in methodology can be addressed by articulating new methods. But when the problem formulation itself is flawed, neither algorithms nor experiments are sufficient to address the underlying problem.'' \cite[p.8]{lipton2018mythos}

Finale Doshi-Velez and Been Kim highlight the problem of assessing the quality of explanations and comparing different explanation techniques. They describe three potential standards for evaluation: application, human, and functionally grounded interpretability, the first two rely on human studies and the third one on formal model properties \cite{doshi2017towards}. They posit the intuitive principle that ``the claim of the research should match the type of the evaluation.'' \cite[p.9]{doshi2017towards}

Cynthia Rudin provides examples of post-hoc explanations that can mislead the user because they are difficult to interpret \cite{rudin2019stop}. She argues that this issue becomes particularly threatening when the stakes are high, and model authorities have a financial interest in model opacity. Rudin and her co-authors point out that: ``interpretable models do not necessarily create or enable trust – they could also enable distrust. They simply allow users to decide whether to trust them.'' \cite[p.6]{rudin2022interpretable} In consequence, they argue in favor of inherently interpretable models.

Our views on XAI have also been strongly shaped by philosophical discussions around explanation and interpretability. Philosophers gave formal accounts of what constitutes an explanatory relationship, namely a statement about the phenomenon to be explained (called the \emph{explanandum}), a statement about a phenomenon that explains the explanandum (called the \emph{explanans}), and an \emph{explanatory link} between explanans and explanandum \cite{sep-scientific-explanation,hempel1948studies}. For formalizing the explanatory link, especially causal accounts dominate, where the explanans is a difference maker with respect to the explanandum \cite{woodward2005making}. 

Krishnan rightfully highlights the importance of distinguishing the causal explanatory from the justificatory role of explanations. She notes that the two may often not align in the context of XAI as we might face explanations that do not justify decisions and justifications that do not explain them \cite{krishnan2020against}. Others have emphasized the different explananda present in XAI, are we interested in explaining the model or the modeled real-world phenomenon \cite{freiesleben2022scientific,sullivan2020understanding,watson2022conceptual}? Finally, Erasmus, Brunet, and Fisher argued that many statements may formally explain a phenomenon, however, it is often difficult to interpret these explanations correctly \cite{erasmus2021interpretability}.


\section{Misconceptions in XAI Research}
In this section, we highlight the key misconceptions we see present in current XAI research and illustrate them in little caricatures. For many of these misconceptions, we are not the first to identify them. However, these misconceptions have persisted over time despite strong and convincing criticism. We see nothing wrong in repeating true things that are still ignored by parts of our community.

\subsection*{Misconception 1: ``Explanation Methods are Purpose-Free''}
Many 'explanation methods' are presented as mathematical constructs without a conceptual or practical justification. Usually, such papers have the following storyline: 
\begin{enumerate}
    \item ML models are black-boxes
     \item Explanations are needed because of [trust, transparency, detecting bugs, etc.]
     \item Here are some formalisms, theorems, and the implementation
     \item Look at the nice [images, text annotations, plots, etc.], don't they look exactly how you would expect them?
     \item In this arbitrary benchmark we invented, our method is much better than all the others in 'explaining'. 
\end{enumerate}
However, it remains unclear why anyone should call these images or plots explanations in the first place. Worse, it even remains unclear what purpose these 'explanations' might serve and under what conditions they are helpful. 

\begin{figure}[h]
\centering
   \includegraphics[width=0.9\linewidth]{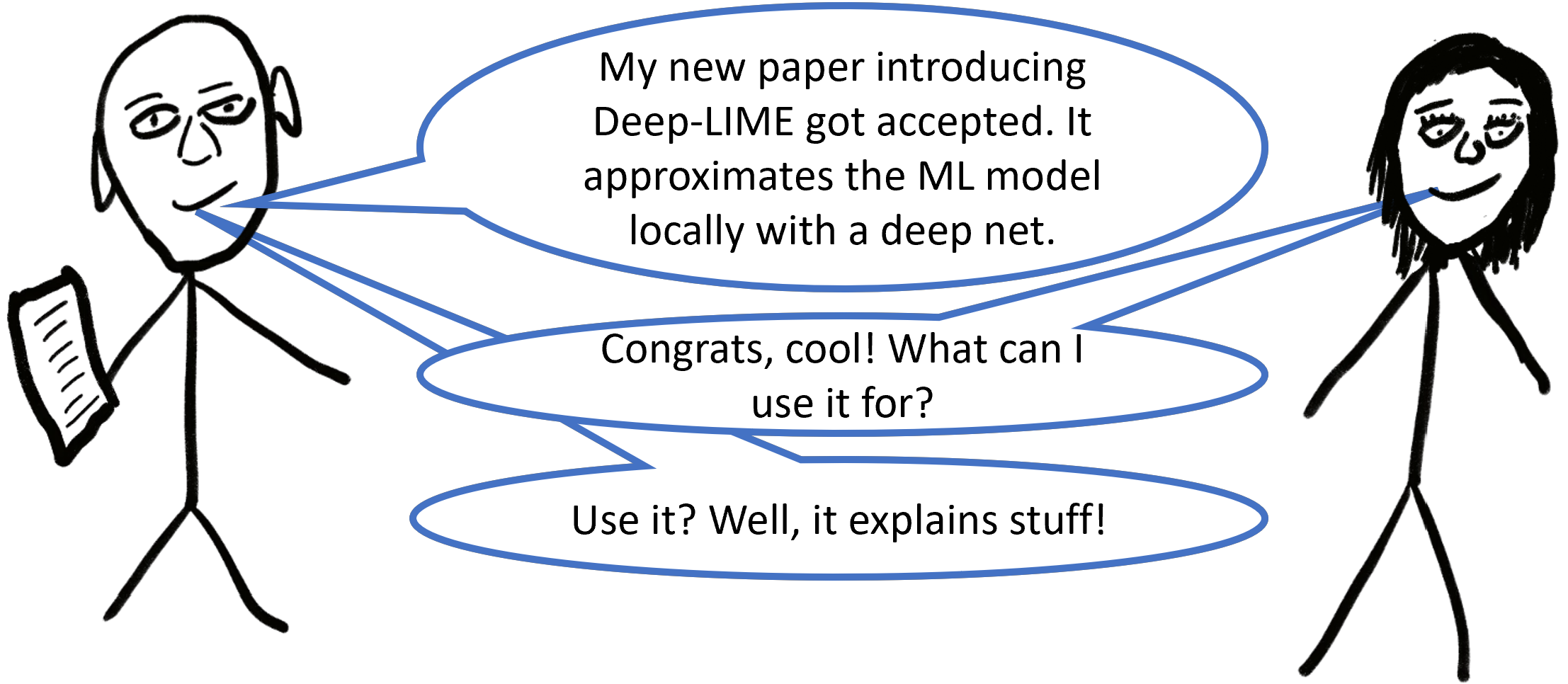}
  \caption{Misconception that explanation methods are purpose-free} 
\label{fig:purpose}
\end{figure} 

We do not claim that explanations can serve only one purpose, but rather that they should serve at least one purpose. Moreover, it should be shown, or at least clearly motivated, how exactly the proposed explanation technique serves this purpose. One may contend here that we do science for science's sake; the purpose is knowledge. However, as long as we do not have a widely accepted definition of explainability or interpretability, a purpose is the only way to connect explainability techniques with the real world. 'Explanation techniques' that are not motivated by any practical purpose should be suspicious to our community. If you cannot think of any context in which your explanation helps potential explainees (i.e. the recipients of explanations), this is a good indication that you should trash the technique.

\subsection*{Misconception 2: ``One Explanation Technique to Rule Them All''}

There is a persistent belief in our community that we only need to find and research the single best explanation technique (e.g., SHAP), choose the best hyperparameters (e.g., the ideal baseline), and then we will always have the best explanations that provide perfect understanding. However, the goals we pursue with explanations are diverse: we may want to audit the model, learn something about the modeled phenomenon, debug models, or provide end-users with the ability to contest the model's decision or act based on it. Depending on the goal, an entirely different technique, with different hyperparameters choices and additional side constraints may be appropriate. 

\begin{figure}[h]
\centering
   \includegraphics[width=0.9\linewidth]{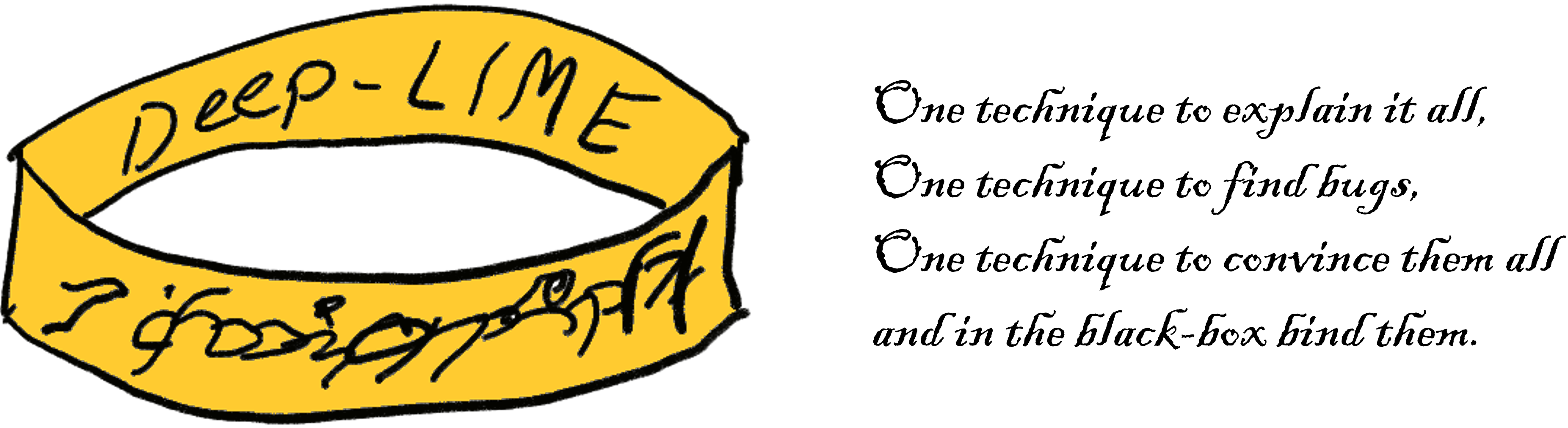}
  \caption{Misconception that there is one true explanation technique} 
\label{fig:one-explanation}
\end{figure} 

Explanation Purposes are generally in conflict. Counterfactual explanations are the ideal example to illustrate these conflicts and the trade-offs we must make \cite{konig2022improvement}. In the original paper by Wachter et al \cite{wachter2017counterfactual}, counterfactuals are presented as explanations that provide understanding, contestability, and recourse. If we think of algorithmic recourse (counterfactuals that guide human actions to reach a desired outcome), the actionability of features is crucial; for example, humans cannot simply become younger to reach the desired outcome. Thus, age is not part of counterfactuals tailored for recourse. Discrimination based on age, on the other side, might be a good reason to contest a decision. That is why, age can surely be part of a counterfactual tailored for contesting. Finally, for the vague purpose of understanding the ML model, counterfactuals might not be the right tool at all, as they only provide extremely limited insight into the model.

\subsection*{Misconception 3: ``Benchmarks do not Need a Ground-Truth''}
Benchmarks are meant to be objective comparisons between competitors according to a universally agreed standard. Machine learners love benchmarks. Benchmarks have been the bread and butter in ML research in the last decade and an important pillar for progressing the field. 
Because of the success of benchmarking in ML, the XAI community figured that benchmarks should be a central part of our field as well. Unfortunately, in XAI we generally lack the central element we have in supervised ML to make objective comparisons -- a ground truth. Without a ground-truth, it is hard to come up with metrics that quantify desirable properties and that are widely agreed upon.

\begin{figure}[h]
\centering
   \includegraphics[width=0.9\linewidth]{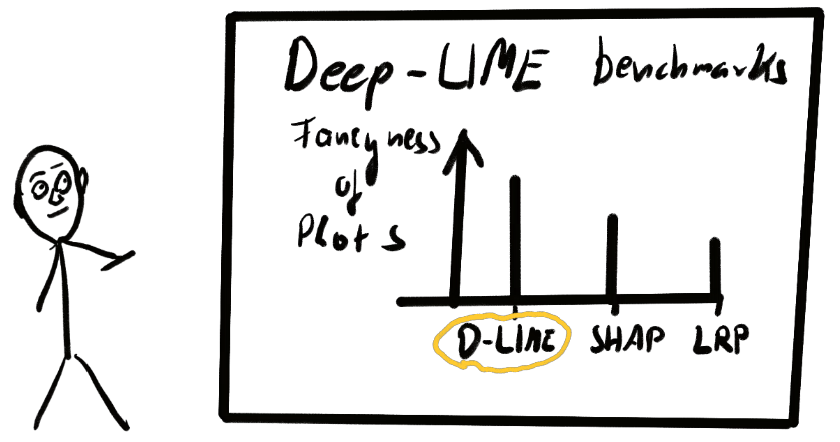}
  \caption{Misconception that we can have benchmarks without ground-truth} 
\label{fig:benchmarks}
\end{figure} 

Accepting the problem of the missing ground truth, there would have been two ways for progress in XAI: 1. abandon the idea of benchmarks in XAI altogether and move toward a more qualitative evaluation of explanations; 2. define benchmarks through the explanation purpose, i.e., how well does the explanation serve that purpose, which gives us again some notion of ground-truth. Parts of our community, however, have taken less rocky paths: Regardless of the explanation purpose, and with little conceptual motivation, they formally define properties that they are optimizing their explanations for. Other explanation techniques (often designed for completely different applications and optimized for distinct desiderata) are then compared according to their own standards. In this form, benchmarks lose their justification; they become advertisement space rather than an objective standard for comparison.

\subsection*{Misconception 4: ``We Should Give People Explanations They Find Intuitive''}
Many papers in our field use standards to motivate explanations that we find particularly questionable. For instance, explainees are given images or annotations that should convince them that the explanation technique actually highlights the right things. The images and annotations are tailored to look compelling and intuitive, conveying a message like -- ``You see, the model is actually looking at the parts of the object that you also look at when performing the task; you can trust this.'' As a consequence, we (over-)fit explanation techniques to human intuition; however, the question is whether these 'explanations' are still faithful to the explained ML model. 



\begin{figure}[h]
\centering
   \includegraphics[width=0.9\linewidth]{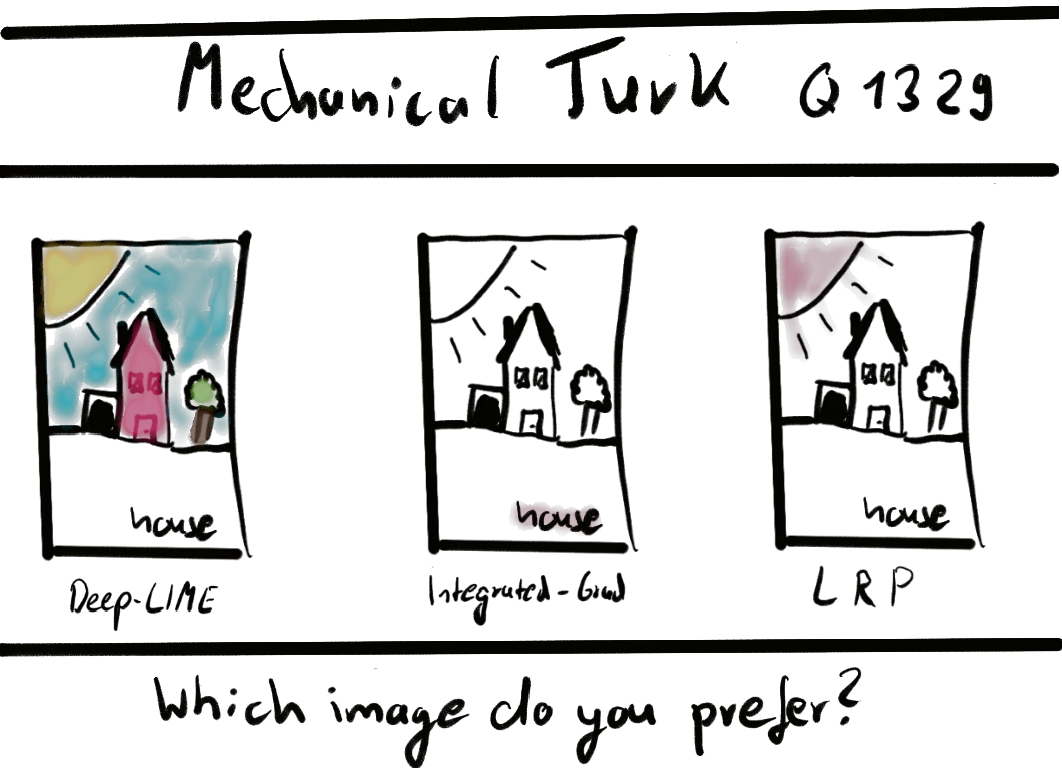}
  \caption{Misconception that the goal is to give people explanations they find intuitive} 
\label{fig:liking}
\end{figure} 

We think that a categorical mistake is made here; XAI should help make the model mechanism more transparent, not compel people into believing the system is good. Explanations provide grounds to decide whether to trust the model; they should not be designed to compel people into trusting the model. We should distinguish between an \emph{explanation} of a decision and a \emph{justification} of a decision. Justifications are good reasons for a decision; Explanations are the actual reasons for a decision \cite{sep-reasons-just-vs-expl,krishnan2020against}. They may align in decisions where the actual reasons for a decision can be ethically justified. In XAI, however, they very often diverge. Think of cases where an 'explainer' 'explains' the predictions of the prediction model without any access to it beyond the single prediction. Or, when the evaluation standard for explanations is which kinds of explanations people like better. Indeed, it can be argued that people also often provide only justifications for their actions, but do not provide their actual reasons or are often not even aware of them. However, this is not an argument for why we should accept the same for XAI explanations; instead, we should strive for higher standards, explanations that are faithful to the causal decision-making process \cite{gunther2022algorithmic}.

\subsection*{Misconception 5: ``Current Deep Nets Accidentally Learn Human Concepts''}

Big parts of our field share the following, in our opinion unwarranted, presupposition: Deep neural nets learn the same concepts as humans.  The idea is that early layers learn low-level concepts, such as edges in images or syllables in sound; Layers closer to the output on the other side learn high-level concepts, such as the concept of a wheel or the concept of a noun \cite{olah2017feature}. Concepts are assumed to be learned without explicitly forcing the model to learn such concepts, but only by optimizing the model to classify images or correctly complete sentences. The assumption is that the only way to solve complex tasks is to use exactly the concepts that humans use \cite{beckmann2023rejecting}. Thus, all we need to do is to train the network and then use XAI techniques like activation maximization or network dissection to discover/reveal which nodes in the network stand for which concept, and then -- tada -- we have a fully transparent model where every part of the model stands for something, and the model basically does logical reasoning again \cite{olah2020zoom}.

We agree that this would be fantastic; however, for the following reasons, we are far more pessimistic concerning the conceptual reasoning in neural nets:

\begin{itemize}
    \item Many regularization techniques, for instance, dropout \cite{srivastava2014dropout}, explicitly force the model to represent in a distributed manner by punishing overreliance on individual neurons.
    \item Even though research showed that some nodes in the network co-activate in the presence of certain concepts (actually, the co-activation in percentage is far less impressive than one would think), the causal role of the concept is not shared \cite{bau2017network,mu2020compositional,voss2021visualizing,donnelly2019interpretability,GALE202060}. That means that for instance cutting the neuron in a bird classifier that 'represents' wings  or intervening on it does not or only marginally change the model's performance/prediction when birds with different wings are presented. Is this really what we mean when we talk about representing concepts?
    \item One of the reasons why humans have shared concepts is because they need to effectively communicate with other humans about the world \cite{grice1989studies,sperber1986relevance,pagel2017q}. However, effective communication has not been a constraint in the training of ML models. Also, humans do not face one but a variety of different tasks. For simple classifications, abstract concepts are not needed as there exist shortcuts \cite{geirhos2020shortcut}.
\end{itemize}

\begin{figure}[h]
\centering
   \includegraphics[width=0.9\linewidth]{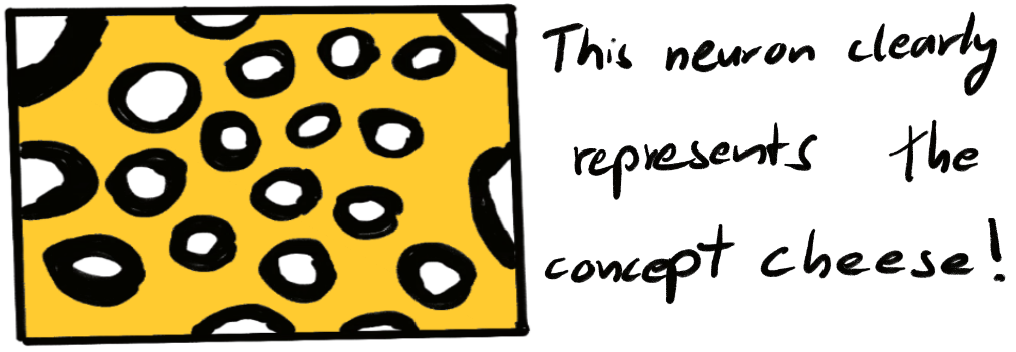}
  \caption{Misconception that current deep nets accidentally learn human concepts} 
\label{fig:concept}
\end{figure} 

Fancy images like those generated by activation maximization techniques \cite{olah2017feature,molnar2020model} should not fool us in this regard: Just because the images generated have some wing-like elements does not mean that they represent wings. Not only are the images we get extremely sensitive to the source image on which we perform activation maximization \cite{olah2017feature}, but they are likely to contain other forms and small shapes that we, as humans, blend out. For instance, research on adversarial examples indicates that deep nets use features in their classification that humans do not attend to \cite{ilyas2019adversarial}. It is questionable whether we as humans will ever understand the 'concepts' of ML models \cite{buckner2020understanding}.

\subsection*{Misconception 6: ``Every XAI Paper Needs Human Studies''}

Many pointed to the importance of human studies in making progress on XAI \cite{liao2021human,doshi2017towards,das2020opportunities}.  We agree that evaluating the quality of explanations based on their impact on human performance on a particular task (to which the explanations are tailored) is reasonable and solid research. However, when it comes to explaining a specific phenomenon, at least two distinct questions must be addressed \cite{miller2019explanation}: 1. What counts conceptually as an explanation for the phenomenon? 2. Which among the explanations for the phenomenon are good explanations for a specific explainee? While the latter question requires properly designed human studies, the former does not; instead, it's a philosophical/conceptual question that can be addressed with conceptual analysis and formal mathematical tools.

Why is the conceptual definition of what counts as an explanation important at all? Why can't we go directly to the second step and test explanations in the real world, with real human explainees? In principle we could do that, but in practice the space of possible 'explanations' is unlimited. Conceptualizing what counts as an explanation for a phenomenon is building up the theory needed for an informed search for good explanations. In many cases where human studies are conducted, a more careful conceptual analysis would have been advisable.

\begin{figure}[h]
\centering
   \includegraphics[width=0.9\linewidth]{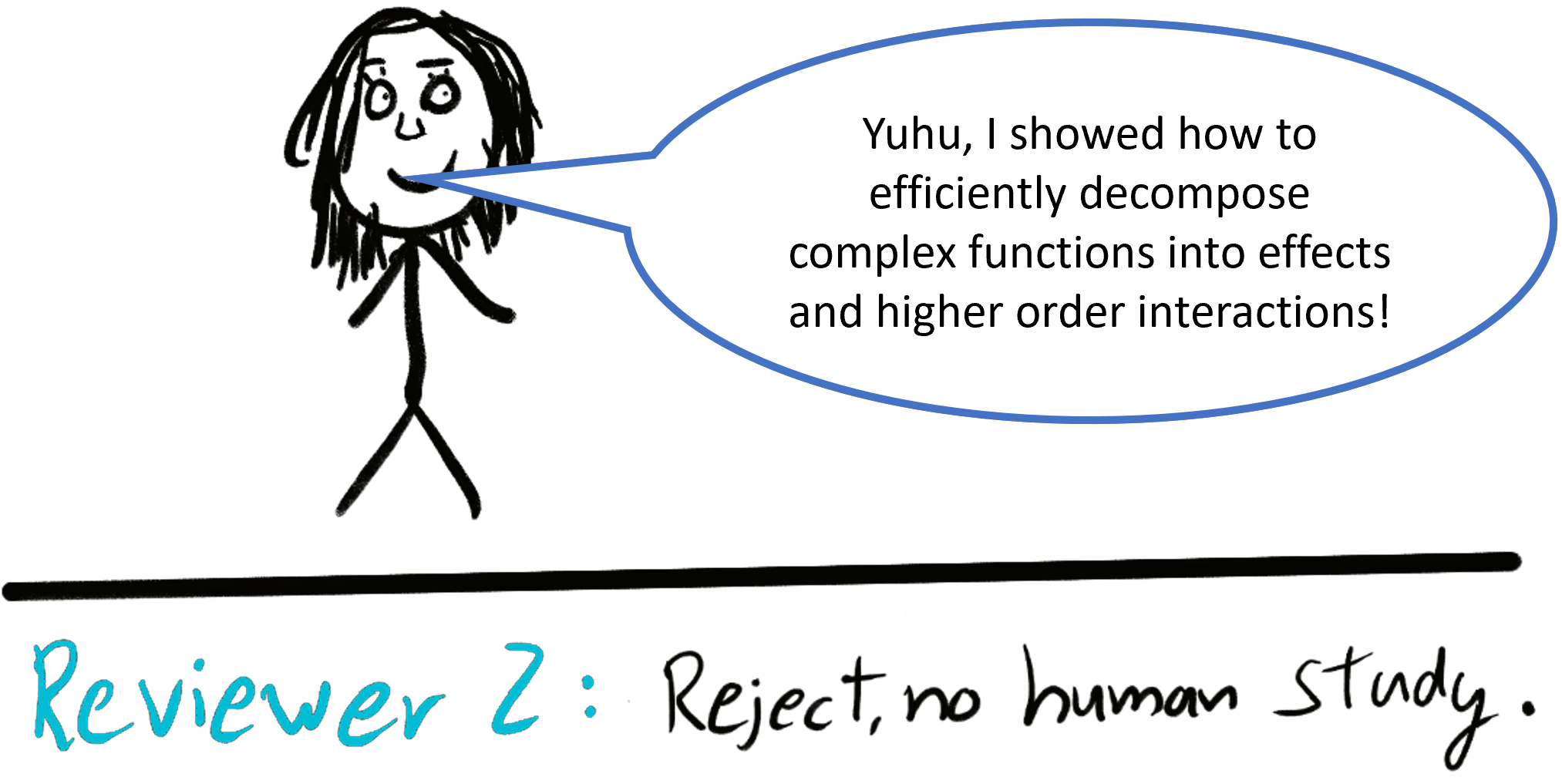}
  \caption{Misconception that every XAI paper needs human studies} 
\label{fig:experiments}
\end{figure} 

More generally, not conducting human studies does not mean dismissing explanation evaluation. For instance, a purely formal evaluation of explanation techniques can be justified if human studies have already been conducted for that type of explanation. Also, not all purposes of XAI require conducting human studies. For example, if we want to use XAI to estimate a specific quantity using the model, the speed and accuracy by which this quantity is measured allows us to compare it with other estimators estimating the same quantity \cite{molnar2021relating}.

\subsection*{Misconception 7: ``XAI Methods can be Wrong''}

Many papers have recently shown how saliency-based or model-agnostic explanation techniques like SHAP, LIME, counterfactuals can be 'tricked' to provide any desired explanation \cite{slack2020fooling,lakkaraju2020fool,adebayo2018sanity,kindermans2019reliability}. This has been taken as major arguments against these techniques and led to arguments why the techniques are wrong or questioning their reliability \cite{lakkaraju2020fool,rudin2019stop,rudin2022interpretable,watson2022conceptual}. To us, there seem to be misunderstandings concerning the consequences of these lines of research.

\begin{figure}[h]
\centering
   \includegraphics[width=0.9\linewidth]{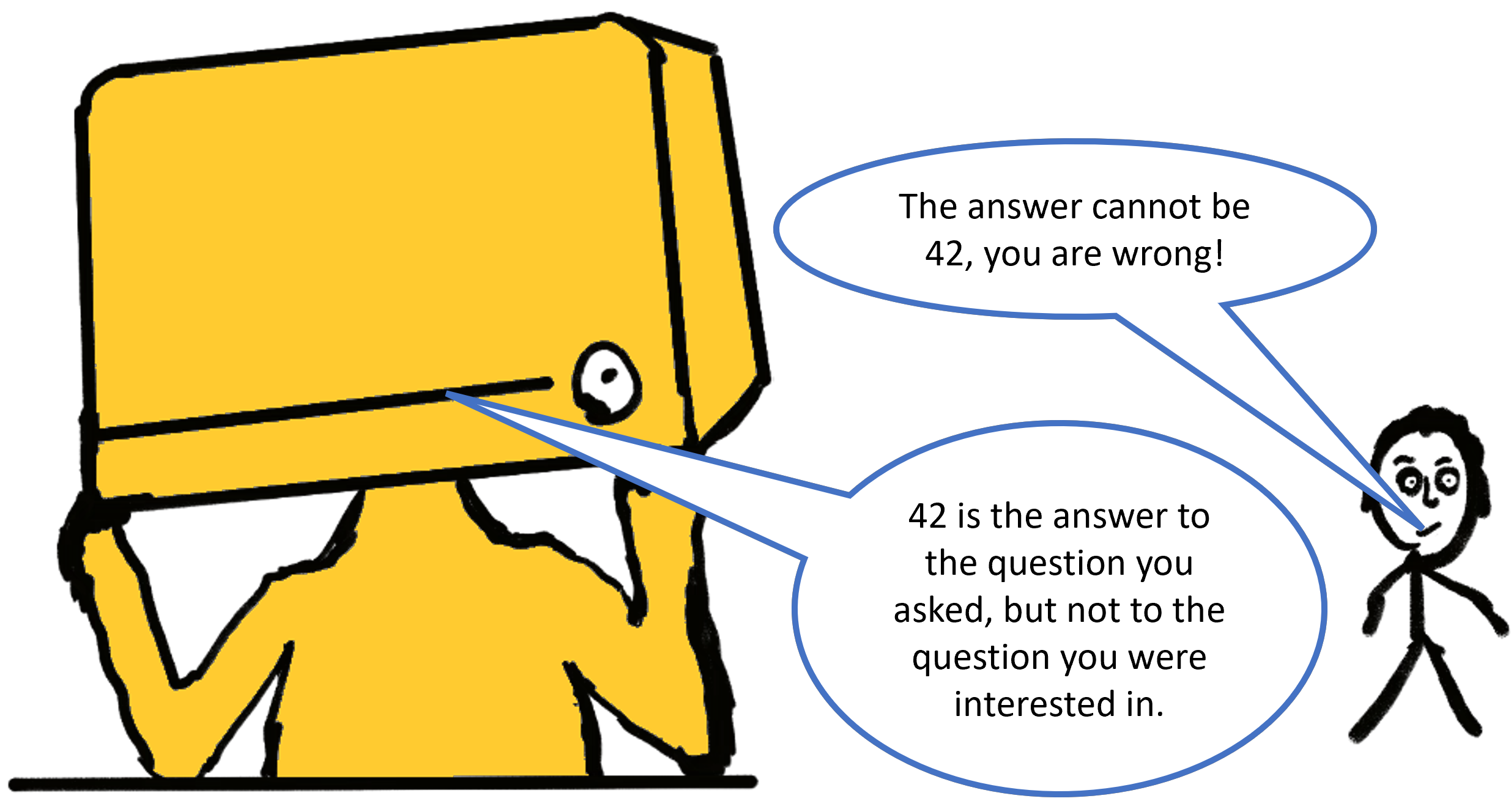}
  \caption{Misconception that XAI methods can be wrong} 
\label{fig:explWrong}
\end{figure} 

While we allow for arbitrary model and data complexity, we require that explanations be simple. Therefore, explanations will indeed not be faithful to every aspect of the model. In this sense, they do nothing wrong; they describe the formal aspects they describe. The fact that explanations are not faithful to every model aspect is the motivation for having different kinds of XAI techniques, each illuminating a different aspect while neglecting another. You may be able to fool SHAP, you may be able to fool LIME, but you won't be able to fool all techniques all the time. It is difficult to find the right level of abstraction in a given context: easily interpretable and local explanations like counterfactuals might have too little expressive power, they can be manipulated without changing much of the overall model behavior; more abstract and global explanations like partial dependence plots may zoom out too far, thereby allowing to hide problematic behavior in the specifics of the model. 

The fact that small model modifications can mislead explanation techniques is nevertheless important -- it shows that the XAI techniques we have and the explanations they provide are very hard to interpret. We may need more diverse evidence to draw conclusions based on XAI explanations. Our field should take this as a call for developing XAI techniques on all levels of abstraction, describing all aspects of behavior relevant to real-world purposes.


\subsection*{Misconception 8: ``Extrapolating to Stay True to the Model''}

Most XAI techniques rely on probing the ML model in one way or the other: LIME is locally sampling inputs, predicts them, and fits a linear model; counterfactuals search for close input points from a desired predicted class; Permutation feature importance (PFI) permutes the values in a specific feature and measures the drop in performance due to this permutation; Activation maximization uses gradient descent to find an input that maximally triggers a specific unit; integrated gradients approximate the integral over the path integral between the 'explained' image and a baseline image.

\begin{figure}[h]
\centering
   \includegraphics[width=0.9\linewidth]{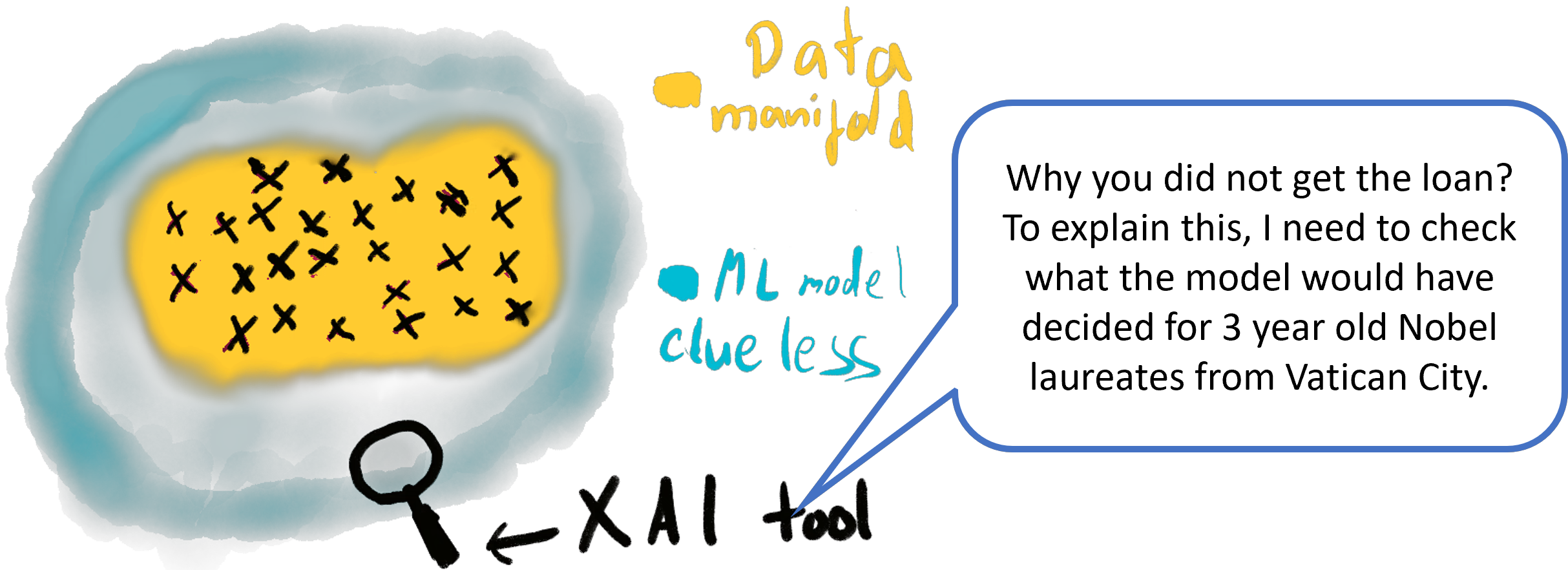}
  \caption{Misconception that model evaluation in extrapolation regions is unproblematic} 
\label{fig:extrapolation}
\end{figure} 

The problem is not THAT the model is probed, but WHERE -- namely in areas where it has not seen any data, i.e., in areas where the model has to extrapolate \cite{hooker2021unrestricted}. ML models are notoriously bad at extrapolating to completely unseen instances \cite{haley1992extrapolation,barnard1992extrapolation,hasson2020direct}. In extrapolation regions, models disagree even when fitted to exactly the same data and achieve similar high performance on a test set. Asking an ML model to extrapolate is like asking a five-year-old kid who hasn't gone to school about her insights into algebraic topology. You might get an answer, but that answer will not really help you. 

Recent literature argues that explanations that rely on extrapolation are true to the model, while those that only probe the model within the data manifold are true to the data \cite{chen2020true}.\footnote{If we stay within the manifold, the model explanations can even be interpreted in terms of the data-generating mechanism \cite{freiesleben2022scientific}.} Clearly, since the model is defined for instances outside of the manifold, probing the model in these areas will give us further insight into the model (for purposes such as debugging or robustness checks) that we would not have gained otherwise. However, we believe that for most XAI purposes, we are interested in the behavior of the model in areas where it is (at least putatively) qualified.  As soon as we leave the data manifold, the interpretation of explanation techniques becomes very blurry. We think it is highly problematic for the interpretation of current explanation methods that they rely so strongly on extrapolation.

\section{Steps Forward}

We hope that these misconceptions show: XAI is still a pre-paradigmatic discipline \cite{kuhn1970structure}. We cannot simply adopt some arbitrary assumptions and move on to paradigmatic scientific problem-solving. We must fight about the right conceptions of what the field is about, the language we should use, and the right evaluation strategies. We know that it is very easy to be critical while it is very difficult to be constructive. So we want to share at least some thoughts and intuitions about how we think the field should evolve to become a more substantive discipline.

\subsection*{Step 1: Go From Purpose to Benchmark}

Explanation techniques should start with a purpose. Again, this does not mean that they can only serve one purpose, but they should show that they serve at least one purpose. A purpose is a goal humans have in mind when they ask for explanations. Once the purpose is fixed, the evaluation of the explanations follows naturally. Your explanation technique should enable debugging? Then the evaluation for the method should be a qualitative study of whether the method suits model developers and helps them to debug their models. If your global explanation technique is supposed to infer relevant properties of the data-generating mechanism, then show in a simulation how well and how resource efficient your technique approximates these properties. When your local explanation technique is designed to provide recourse options to end-users, then either carefully conceptually justify desiderata for recourse and base your evaluation upon these desiderata, or test the suitability of these recourse options in experiments.

The purpose determines the right evaluation metric; the evaluation metric(s) often allows for benchmarking. Different explanation techniques that are designed for the same purpose can be judged by the same evaluation metric(s) and thus benchmarked. One simple example is when two methods are estimating the same quantity i.e. a quantifiable property of the model. 

\subsection*{Step 2: Be Clear What You Need to Explain and by What}
Every explanation comes with an explanation target, the so-called \emph{explanandum}. The explanandum specifies what is to be explained and is determined by the explanation's purpose. Very often, confusion in XAI research arises because it is unclear what the explanandum is in a given context. For instance, confusions about the right sampling technique are often implicit confusions about the right explanandum \cite{watson2022conceptual,freiesleben2022scientific}. XAI techniques may for instance aim to explain:
\begin{itemize}
    \item the model prediction $\hat{Y}$,
    \item the predicted target $Y$, or
    \item an intermediate model element.
\end{itemize}

If you are clear about the explanandum, the second big question is by what you want to explain it -- the so-called \emph{explanans}. The explanans describes the factor(s) you are pointing to in order to account for the state of the explanandum. There are a variety of explanantia (plural of explanans) in XAI research such as:
\begin{itemize}
    \item the model inputs $\overline{X}$,
    \item the predictors $X$,
    \item the dataset or a subset of it, or
    \item intermediate model elements.
\end{itemize}

Finally, be clear on the connection between the explanans and the explanandum. Explanations can be established by pointing to associations between the explanans and the explanandum \cite{salmon1971statistical}. Usually, however, the relationship we are interested in is causal, that is, the explanans makes a difference for the explanandum \cite{woodward2005making}. While causal explanations are more desirable than reference to mere associations, they are also more difficult to establish.

\subsection*{Step 3: Give Clear Instructions for How to Interpret Explanation Techniques}
Interpreting the outputs of XAI techniques is extremely difficult. Rather than letting people figure out how to interpret XAI statements on their own, papers should provide clear guidance on how to do so. We believe that addressing the following questions in new proposals for XAI techniques would contribute to securing good usage:

\begin{itemize}
    \item What purpose does this XAI technique serve and how should it be applied?
    \item Under which (model) conditions does the XAI technique enable a clear interpretation? 
    \item How do the hyperparameters of the technique affect the interpretation?
    \item What is the intuitive meaning of extremes, namely high, close to zero, or negative values? 
    \item In what way, does the explanation guide actions and decisions?
    \item When is it better to rely on other explanation techniques and why? 
\end{itemize}


\subsection*{Step 4: XAI Needs Interdisciplinarity AND Expertise}

XAI is a highly interdisciplinary field. XAI involves so many aspects that a single field would fail terribly; we need interaction. XAI needs to solve the following key questions, among others:

\begin{itemize}
    \item \textbf{Conceptual:} What are relevant explanation purposes? What is required to establish an explanatory relationship between an explanans and an explanandum? What are general explanation desiderata for a specific purpose? How can explanations be conceptualized? How to interpret explanations?
    \item \textbf{Technical:} How to describe the conceptual definitions formally? What can be shown formally about the properties of these explanations? How to compute the explanations efficiently? How to implement explanations accessibly and correctly? How to interpret formalized explanations?
    \item \textbf{Psychological:} How to visualize explanations the right way? What makes a good explanation for a particular explainee? What are context and person-specific desiderata of explanations? What cognitive biases do people have when interpreting explanations? Is the explanation successful in serving the explanation purpose?
    \item \textbf{Social and Ethical:} Should we provide explanations and if yes, what are ethical desiderata? What are the risks with XAI in high-stakes decisions? How do explanations affect people's trust and actions? What level of transparency do we need?
\end{itemize}

Not every paper must involve researchers from each group. However, the questions between the different categories should be seen as closely tied: Formal XAI methods without a conceptual foundation should be disregarded; Conceptually solid XAI tools that experimentally fail in guiding humans should be modified and fine-tuned; Finally, XAI explanations that serve a purpose successfully but this purpose is morally questionable should be dismissed.

At the same time, nothing is wrong with XAI research that focuses on a narrow field-specific question such as providing a more efficient algorithm or testing a specific XAI method in human experiments concerning its success in finding model flaws. Every field has its expertise and it is important that conceptual foundations, algorithms, experiments, and ethical evaluations live up to the highest standards of the individual fields. All we want to emphasize is to not run around having blenders on but recognize how the questions are intertwined.

\section{Conclusion}
This paper covered the key misconceptions in current XAI research. In our opinion, the most important one is the idea of purpose-free explanations. Fixing specific purposes will provide a way for evaluating and benchmarking XAI techniques objectively. The explanation purpose will also guide us: how XAI techniques must be constructed, when they should be used, and how they have to be interpreted. Overall, purpose-centered XAI research will help us make ML systems more transparent. Therefore, we hope that future researchers will start thinking more about the purpose of explanations before they make grand proposals for new methods.

\section*{Acknowledgements}

This project has been supported by the German Federal Ministry of Education and Research (BMBF) and the Carl Zeiss Foundation (project on ``Certification and Foundations of Safe Machine Learning Systems in
Healthcare'').

%
%
\newpage
 \bibliographystyle{splncs04}
%
\bibliography{biblio.bib}
\end{document}